\documentclass[conference]{IEEEtran}
\usepackage{cite}
\usepackage{epsfig} 
\usepackage{amsmath,amssymb,amsfonts}
\usepackage{algorithmic}
\usepackage{graphicx}
\usepackage{textcomp}
\usepackage{hyperref}
\usepackage{makecell}
\usepackage{multicol}
\usepackage{xcolor}
\usepackage{multirow}
\usepackage{float}
\usepackage{subcaption}
\def\BibTeX{{\rm B\kern-.05em{\sc i\kern-.025em b}\kern-.08em
    T\kern-.1667em\lower.7ex\hbox{E}\kern-.125emX}}
\begin{document}

\title{Fitness-for-Duty Classification using Temporal Sequences of Iris Periocular images \\

\thanks{This work is supported by the German Federal Ministry of Education and Research and the Hessian Ministry of Higher Education, Research, Science and the Arts within their joint support of the National Research Center for Applied Cybersecurity ATHENE and TOC Biometrics.}
}

\author{
\IEEEauthorblockN{Pamela C. Zurita}
\IEEEauthorblockA{\textit{R\&D Center} \\
\textit{TOC R\&D Center}\\
Santiago, Chile \\
pamela.zurita@tocbiometrics.com}
\and
\IEEEauthorblockN{Daniel P. Benalcazar}
\IEEEauthorblockA{\textit{DIMEC} \\
\textit{Universidad de Chile}\\
Santiago, Chile \\
dbenalcazar@ug.uchile.cl}
\and
\IEEEauthorblockN{Juan E. Tapia}
\IEEEauthorblockA{\textit{Hochschule Darmstadt} \\
\textit{{da/sec-Biometrics and Internet Security Research Group}}\\
Darmstadt, Germany \\
juan.tapia-farias@h-da.de}
}


\IEEEoverridecommandlockouts
\IEEEpubid{\makebox[\columnwidth]{979-8-3503-3607-8/23/\$31.00 ©2023 IEEE \hfill} \hspace{\columnsep}\makebox[\columnwidth]{ }}

\maketitle

\IEEEpubidadjcol


\begin{abstract}
Fitness for Duty (FFD) techniques detects whether a subject is Fit to perform their work safely, which means no reduced alertness condition and security, or if they are Unfit, which means alertness condition reduced by sleepiness or consumption of alcohol and drugs. Human iris behaviour provides valuable information to predict FFD since pupil and iris movements are controlled by the central nervous system and are influenced by illumination, fatigue, alcohol, and drugs. This work aims to classify FFD using sequences of 8 iris images and to extract spatial and temporal information using Convolutional Neural Networks (CNN) and Long Short Term Memory Networks (LSTM). 
Our results achieved a precision of 81.4\% and 96.9\% for the prediction of Fit and Unfit subjects, respectively. The results also show that it is possible to determine if a subject is under alcohol, drug, and sleepiness conditions. Sleepiness can be identified as the most difficult condition to be determined. This system opens a different insight into iris biometric applications.

\end{abstract}

\renewcommand\IEEEkeywordsname{Keywords}


\section{Introduction}

The iris biometric can be highlighted as one the most accurate biometric modalities, with a great expansion margin in the next years. Features such as contactless capture, no invasive, faster iris identification, and lower capture device prices are relevant for developing complementary biometrics applications on biometrics \cite{Jain, bowyer2016handbook}. These properties, along with more availability of cost-effective capture devices, allow the development of new biometric applications to complement iris recognition. A complementary insight is determining Fitness for Duty (FFD) using Near Infrared Range (NIR) iris images. 

FFD systems aim to detect if a person is capable or not (Fit/Unfit) of carrying out their daily tasks or if they are impaired by fatigue/sleepiness or alcohol/drug consumption \cite{macquarrie2018fit, murphy1992fitness}. Detecting such impairments is very important because it saves lives and avoids productivity loss, work accidents caused by operating heavy machinery and transporting people, and even healthcare negligence \cite{causa2022ffd, tapia2022ffd, tapia2022alcohol}. For instance, between 5\% and 40\% of injuries presented at emergency departments in hospitals of 27 countries are due to alcohol consumption \cite{chikritzhs2021alcohol}.

Traditionally, FFD was detected using tests such as psycho-motor tasks, finger tapping, pattern comprehension, smart band wrist, and in-cab monitoring \cite{miller1996fit, serra2007criteria}. However, those tests can take a long time and are subject to impersonation because they do not consider a biometric modality \cite{tapia2022ffd}. 

Today, we have a worldwide dependency rise in alcohol consumption and drugs in the workforce, especially among shift-worker. Europe is not far away from this problem \footnote{\url{https://www.emcdda.europa.eu/publications/data-fact-sheets/european-web-survey-drugs-2021-top-level-findings-eu-21-switzerland_en}}. 
Today, many companies include drug-place-free protocols to care for workers and save lives. Some professionals are more relevant to be measured, such as doctors to avoid negligence in surgery because of alcohol or drug consumption, truck drivers transporting explosive, chemical and other dangerous cargo, commercial pilots because of drugs and also insurance companies. Companies should pay for accidents, not for negligencies. 


It has been demonstrated that there is an important change in iris behaviour depending on whether the subject has consumed alcohol or drugs or is sleepy. It is possible to detect iris changes through NIR images, helping the detection of FFD. \cite{causa2022ffd}\cite{tapia2022ffd}.
The success of these iris-based FFD methods is due to the fact that the Central Nervous System controls the human iris; therefore, iris and pupil movements are involuntary, only affected by factors such as illumination, fatigue and ingestion of psychotropic substances \cite{causa2022ffd, tapia2022ffd, tapia2022alcohol}. It is essential to highlight that this kind of system has no relation with alcohol or drug tests on blood. FFD measure the behavioural changes in the eyes because of external factors.

In this work, we propose to automatically extract spatial and temporal features from a stream of NIR iris by means of Convolutional Neural Networks (CNN) \cite{simonyan2014vgg} in cascade with Long Short Term Memory Networks (LSTM) \cite{Puviyarasu2020LSTMN}. This is an improvement over previous methods that extracted spatial features only \cite{tapia2022ffd, tapia2022alcohol}, or handcrafted temporal features \cite{causa2022ffd}.

The contributions of this work are the following:
\begin{itemize}
\item \textit{Complementary-application}: This proposed system shows that it is possible to use regular NIR capture devices for new applications, such as FFD. 
\item \textit{Sequence Formation}: As a extension for feature extraction, iris-time-sequences information is also included and thoughtfully described. 
\item \textit{Architecture}: A novel and lightweight architecture based on two steps CNN and LSTM are used to extract spatial and temporal features from NIR iris images automatically.
\item \textit{Performance Evaluation}: FFD performance is evaluated and compared against that of state-of-the-art methods.
\end{itemize}

The rest of the article is organised as follows: Section~\ref{sec:relate} summarises the related works on FFD and LSTM. The database description is explained in Section~\ref{sec:dataset}. The metrics are explained in Section~\ref{sec:metrics}. The experiment and results framework is then presented in Section~\ref{sec:results}. We conclude the article in Section~\ref{sec:conclusion}.


\section{Related work}
\label{sec:relate}

Researchers have shown that the best method in order to obtain the best predictions of image frames is through the use of CNN, Recurrent Neural Networks (RNN), LSTM, and a combination of them. In this section, we expand on the most important contributions to feature extraction with a combination of CNN and LSTM and the state-of-the-art in FFD.

\subsection{Long Short Term Memory Networks}
Regarding LSTM, several application has been proposed in state of the art. Sungwoo Jun\cite{SungwooJun8708878}, proposed to use batch-normalisation with an auxiliary classifier that was able to converge faster and had good results in classifying skeleton-based action recognition problems. 

Huanhou Xiao\cite{XIAO2020173} was able to obtain fine-grained captions for Microsoft Video Description Dataset (MSVD) and MSR-Video to Text Dataset (MSR-TT). The proposal was to feature an LSTM structure and dual-stage loss to translate videos into sentences, a task that must consider time series to be well done. He used CNN to generate diverse and fine-grained descriptions and a novel performance evaluation LSTM to asses the fine-grained captions.

Bowen Wang \cite{9382986} created a Convolutional LSTM (ConvLSTM) network that could leverage the temporal coherence in video frames to improve temporal awareness. It replaces a frame in a given video sequence with noises. The training strategy spoils the temporal coherence in video frames, and thus, makes the temporal links in ConvLSTM unreliable. Then improves the ability of the model to extract features from video frames, and serves as a regulariser to avoid overfitting without requiring extra data annotations or computational costs. The implementation is done in CityScapes and EndoVis2018 datasets, city driving videos.

Wijayakoon \cite{article}, handled the colourisation of black and white videos problem by using CNN and LSTM. It is composed of 4 stages: A time-distributed CNN encoder, a time-distributed CNN decoder, a fusion layer, a high-level feature extraction (using Inception ResNet-v2), and an LSTM to extract temporal features within frames.
The combination of CNN LSTM neural networks has shown powerful achievements in the task of classification in different areas, such as autonomous driving, detection of emotions, and human activities.



Jeyanthi \cite{JEYANTHISURESH2020} was able to recognize human actions based on  Inception ResNet, a CNN network, and LSTM. The CNN network is used to extract human features, and LSTM adopts the generic features from the pre-trained CNN. Using two datasets of diverse activities, on UCI 101 and HMDB 51 datasets and a CNN model composed by: VGG-16, Inception-v3, Inception ResNet-v2, and a ResNet-152 network. For this research, Jeyanthi used an LSTM classifier composed of 100 fully connected neurons and a ReLu activation.

Chih-Yao \cite{XIAO2020173} proposed recognising diverse human activity by using the CNN-LSTM structure. First, they demonstrate the strong use of ResNet-101. They applied this baseline thoroughly examine the use of RNNs and Temporal-ConvNets for extracting spatiotemporal information: temporal segment RNN and Inception-style Temporal-ConNet. Each approach requires proper care to achieve state-of-the-art performance: LSTM requires pre-segmented data. 

\subsection{Fitness for Duty}

Regarding FFD, it is a very new topic in the biometric area. Recently, Tapia et al. \cite{tapia2022ffd, tapia2022alcohol} have developed FFD systems based on MobileNetV2 that operate on a single NIR iris image. Those systems reached a good performance using aggressive data augmentation. Capturing an extensive dataset of NIR iris sequences with control, alcohol, drug and sleepiness subjects was a major contribution of those works \cite{tapia2022ffd, tapia2022alcohol}.
The same author proposes a new model based on the fusion of Capsule Network to classify if one subject is under alcohol consumption using a stream of images \cite{ICRP-Tapia}. 

Causa et al. \cite{causa2022ffd} used a different approach. They studied extracting 50 handcrafted features from NIR iris time-sequences as predictors for FFD, using traditional Machine Learning (ML) approaches. To estimate these 50 features, it was necessary to implement an eye detector and segment the iris, pupil and sclera frame by frame. Although ML methods are light, feature extraction was a demanding task, which increased the required computational resources.

Drowsiness has also been an essential topic in FFD. Some research has shown that it can be detected in real-time through an ECG, as Babeian et al. \cite{Babaeian} suggested, but it can be invasive for the user and complex for the person applying it. It can also be detected through iris images. Chellapa et al. \cite{chellapa-Rasp} determined drowsiness through iris images; nonetheless, the performance was not reasonable under low light conditions. This research can only detect drowsiness but not define its cause, such as the consumption of alcohol or drugs or lack of sleep. Some of them also are created for situations where the user is already performing their job, which can be prejudicial. 

\begin{figure*}[]
\includegraphics[width=1\linewidth]{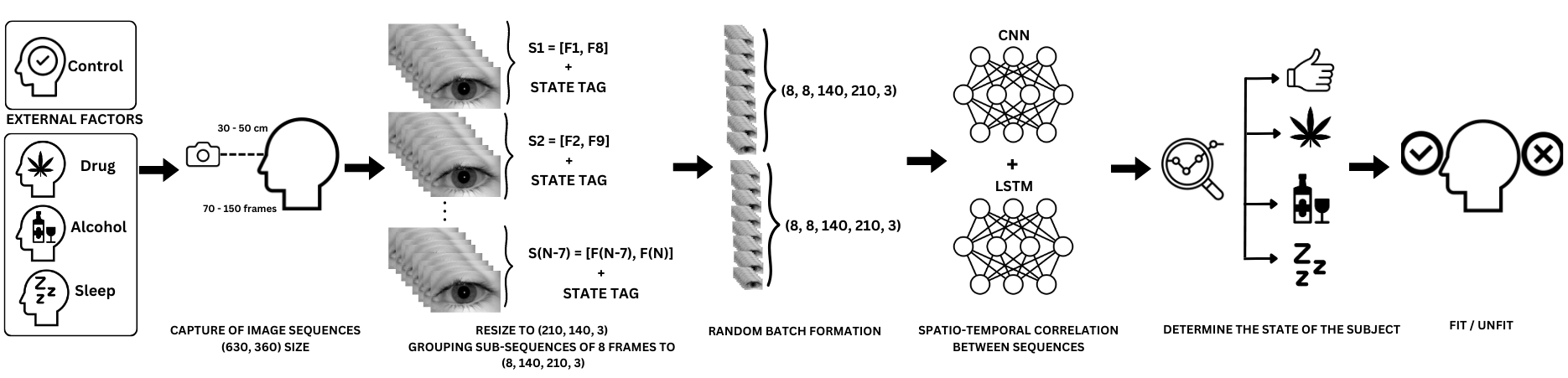}
\caption{Block diagram of the proposed CNN-LSTM model based on periocular eye images. The subject, which can be in an unknown condition for the system, is standing in front of the capture device. Several NIR frames are captured and used as input. The model will extract the features and define the Spatio-temporal dependencies between them to infer whether the subject is Fit or Unfit.}
\label{fig:block}
\end{figure*}

In this work, we improved the previous approaches as a means to automate feature extraction and included the temporal information instead of extracting the 50 handcrafted features. Our proposed features are extracted by a CNN, which extracts the most relevant spatial features from NIR iris frames, followed by an LSTM that analyses the temporal component of those features. Therefore, a robust system automatically extracts temporal and spatial features from iris sequences, as illustrated in Figure \ref{fig:block}.

\section{Metrics}
\label{sec:metrics}
The False Positive Rate (FPR) and False Negative Rate (FNR) were reported as Error Type I and Error Type II. These metrics effectively measure to what degree the algorithm confuses presentations of Fit and Unfit images with alcohol, drugs and sleepness. The FPR and FNR are dependent on a decision threshold. Also, the Confusion Matrix (CM) was reported. The CM is a tabular representation of the performance of an algorithm. Its rows represent the well-predicted class, whereas the columns represent the wrong-predicted classes of alcohol, control, drug and sleep.

A Detection Error Trade-off (DET) curve is also reported for the most difficult class. In the DET curve, the Equal Error Rate (EER) value represents the trade-off when the FPR and FNR. Values in this curve are presented as percentages. Additionally, three different operational points are reported. FNR$_{10}$, which corresponds to the FPR fixed at 10\%, FNR$_{20}$, when the FPR is fixed at 5\%, and FNR$_{100}$, when the FPR is fixed at 1\%. EER, FNR$_{10}$, FNR$_{20}$, and FNR$_{100}$ are independent of decision thresholds.


\section{Method}
\label{sec:method}

\subsection{Dataset}
\label{sec:dataset}

For this research, the "FFD NIR iris images Sequences database" (FFD-NIR-Seq) \cite{causa2022ffd} was used. All the subjects that participated in this database were volunteers, and the capture protocol was approved by Universidad de Chile's health committee. 
This database is composed of sets of up to 10-second stream sequences of periocular NIR images captured at 15 fps. Most subjects have 100 frames available, but the number of frames varies between 75 and 150.

\begin{figure}[t]
\centering
\includegraphics[scale=0.30]{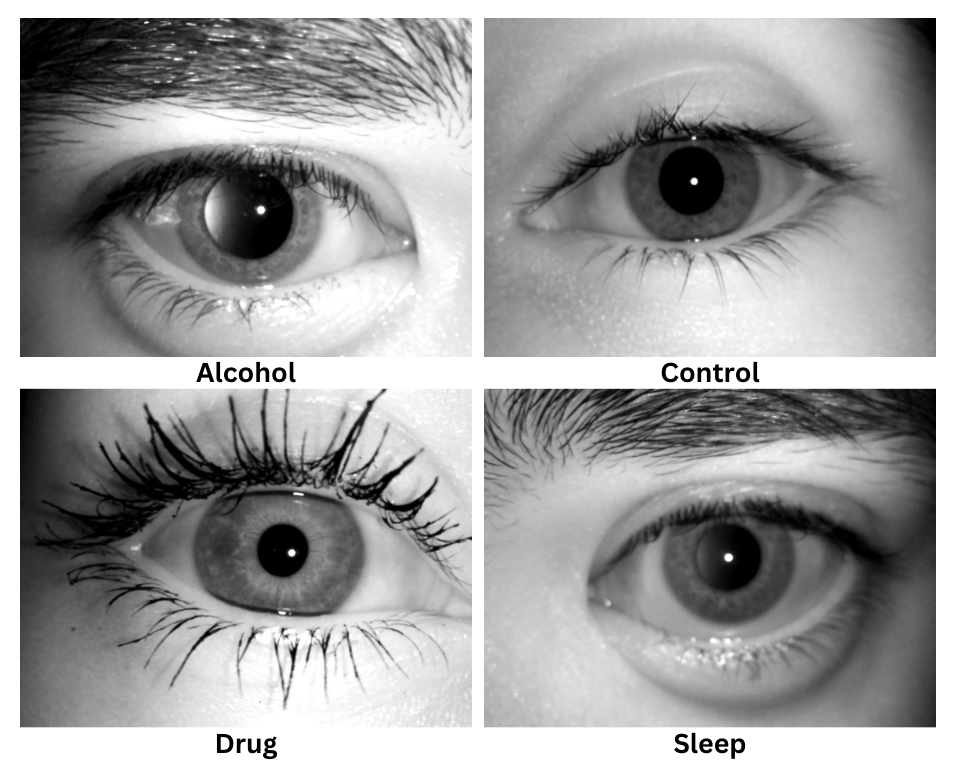}
\caption{Eye-image examples of FFD-NIR-Seq database.}
\label{fig:example_state}
\end{figure}

The binocular NIR image sequences correspond to the periocular area, containing eyes, pupils, iris, and sclera. The size of each image in each sequence is $630 \times 360$ pixels. The number of subjects in the total of this dataset is 980 people, divided as Table \ref{tab:subjects} describes. Each frame's left and right eyes were segmented and cropped using UNet\_xxs as described in \cite{benalcazar2022hardware}. Examples of images captured by the iris sensor are depicted in Figure \ref{fig:example_state}.


The database has four classes of NIR image sequences in different conditions as follows:
\begin{itemize}
   \item 
Control: healthy subjects that are not under alcohol and/or drug influence and in normal sleeping conditions.
   \item 
Alcohol: subjects who have consumed alcohol or are in an inebriation state.
   \item 
Drugs: subjects who consumed some drugs (mainly marijuana) or psychotropic drugs (by medical prescription).    
   \item 
Sleep: subjects with sleep deprivation, resulting in fatigue and/or drowsiness due to sleep disorders related to occupational factors (shift structures with high turnover).
\end{itemize}

As a first step, all the single-eye images were resized to a $210 \times 140$ resolution. Since this work requires sequences of images for identification, consecutive frames were grouped in sub-sequences of 8 frames in the following manner.

For each subject, we have around $N \approx 100$ frames available, namely $F_k$, with $k$ between 1 and $N$. To make the first sub-sequence $S_1$ the first 8 frames $[F_1 , F_8]$ are grouped in a 4D tensor of dimension $8 \times 210 \times 140 \times 3$. Then, to make the next sub-sequence $S_2$, frames $[F_2 , F_9]$ are considered, and so forth. The last sub-sequence for each subject is $S_{N-7}$ that takes frames $[F_{N-7} , F_N]$. Each of these sub-sequences $[S_1 , S_{N-7}]$ were tagged with the subject's state: alcohol, drug, sleepiness, or control. Table \ref{tab:sequences} shows the total number of sub-sequences of 8 frames used in this research. 

\begin{table}[b]
\caption{Dataset by Subjects.}

\centering
    \begin{tabular}{c c c c c}
    \hline
    \hline
        State & Test & Train & Validation & Total \\ 
    \hline
        Alcohol &  72 & 247 & 35 & 354 \\ 
        Control & 688 & 247 &  9 & 944 \\ 
        Drug    &  17 &  62 &  9 &  88 \\ 
        Sleep   &  20 &  69 & 88 & 177   \\ 
    \hline
    \hline
    \end{tabular}
\label{tab:subjects}
\end{table}

\begin{table}[b]
\caption{Dataset by Sequences.}
\centering
    \begin{tabular}{c c c c c}
    \hline
    \hline
        State & Test & Train & Validation & Total \\ 
    \hline
        Alcohol &   6,422 & 22,349 & 3,114 & 31,885 \\ 
        Control &  55,118 & 19,573 & 2,856 & 77,547 \\ 
        Drug    &   2,202 &  8,157 & 1,181 & 11,540 \\ 
        Sleep   &   2,375 &  8,140 & 1,068 & 11,583 \\ 
    \hline
    \hline
    \end{tabular}
\label{tab:sequences}
\end{table}

Finally, a batches dataset containing 8 sub-sequences was performed off-line. This was an efficient way of mixing the batches and storing them in the required resolution \cite{Puviyarasu2020LSTMN}. Additionally, this decreases the time of reading the batches from the disk during training \cite{Puviyarasu2020LSTMN}. To obtain the batches, groups of 8 sub-sequences were chosen from the database in Table \ref{tab:sequences}, forming 5D tensors of dimension $8 \times 8 \times 140 \times 210 \times 3$. The targets of each batch are the corresponding labels of each sub-sequence in the shape of an $8 \times 1$ tensor. The batches contain sub-sequences from the 4 classes: control, alcohol, drugs and sleep.

\subsection{CNN-LSTM Model}
\label{sec:model}

The proposed method is based on \cite{Puviyarasu2020LSTMN}. It is composed of two main modules, a VGG16-inspired CNN and an LSTM. This network receives the labelled $5D$ tensor as an input to the CNN. 
Table \ref{tab:VGG} shows the number of layers, dimensions and parameters of the purpose network. Its output is connected in cascade to the input of the LSTM module, which is described in \ref{tab:LSTM}. 

The model determines the spatio-temporal correlation between the sequences of images to classify them in a certain state: alcohol, control, drug, or sleep. The optimiser used was Adam, with categorical cross-entropy loss. The number of hidden units for the LSTM module was $32$.

The model was created using TensorFlow libraries. The best checkpoints were saved according to the minimum validation loss. The learning rate was set to $ 1^{e-6} $. The Batch Size (BS) was 24; thus it uses 3 of the preformed batches. There was no implementation of data augmentation in this work since available libraries would yield a different augmentation to each image in the sub-sequence; thus, the temporal coherence would be removed.

\begin{table}[htbp]
\scriptsize
\caption{CNN Model Based on VGG-16. BS: Batch Size}
\centering
    \begin{tabular}{c c c c}
    \hline
    \hline
        No & Layer (type) & Output Shape & Parameters \\ 
    \hline
        1 &  InputLayer          & (BS, 140, 210, 3)   & 0  \\ 
        2 &  Conv2D              & (BS, 140, 210, 64)  & 1792 \\ 
        3 &  BatchNormalization  & (BS, 140, 210, 64)  & 256  \\ 
        4 &  MaxPool2D           & (BS, 70, 105, 64)   & 0  \\ 
        5 &  Conv2D              &  (BS, 70, 105, 128) & 73856    \\
        6 &  BatchNormalization  &  (BS, 70, 105, 128) & 512    \\
        7 &  MaxPool2D           &  (BS, 35, 53, 128)  & 0 \\
        8 &  Conv2D              &  (BS, 35, 53, 256)  & 295168    \\
        9 &  Conv2D              &  (BS, 35, 53, 256)  & 590080    \\
        10 &  BatchNormalization &  (BS, 35, 53, 256) & 1024    \\
        11 &  MaxPool2D          &  (BS, 18, 27, 256) & 0 \\
        12 &  Conv2D             &  (BS, 18, 27, 512) & 1180160    \\
        13 &  Conv2D             &  (BS, 18, 27, 512) & 2359808 \\
        14 &  Conv2D             &  (BS, 18, 27, 512) & 2359808    \\
        15 &  BatchNormalization &  (BS, 18, 27, 512) & 2048    \\
        16 &  MaxPool2D          &  (BS, 9, 14, 512)  & 0  \\
        17 &  Flatten            &  (BS, 64512)       & 0 \\
    \hline
    \hline
    \end{tabular}
\label{tab:VGG}
\end{table}

\begin{table}[htbp]
\scriptsize
\caption{LSTM Set-up. BS: Batch Size.}
\centering
    \begin{tabular}{c c c c }
    \hline
    \hline
        No & Layer (type) & Output Shape & Parameters  \\ 
    \hline
        1 &  LSTM               & (BS, 32)    & 8261760 \\ 
        2 &  LSTM               & (BS, 32)    & 8320 \\ 
        3 &  Dense              &  (BS, 1024) & 33792 \\ 
        4 &  BatchNormalization &  (BS, 1024) & 4096    \\
        5 &  Dropout            &  (BS, 1024) & 0 \\
        6 &  Dense              &  (BS, 512)  & 524800    \\
        7 &  Dropout            &  (BS, 512)  & 0    \\
        8 &  Dense              &  (BS, 64)   & 32832    \\
        9 &  Dropout            &  (BS, 64)   & 0    \\
        10 &  Dense             &  (BS, 4)    & 130    \\ 
    \hline
    \hline
    \end{tabular}
\label{tab:LSTM}
\end{table}


\section{Results}
\label{sec:results}

\begin{figure*}[t!]
    \begin{center}
        \subfloat[Binary Confusion Matrix \label{subfig:binary}]{{\includegraphics[width=0.32\linewidth]{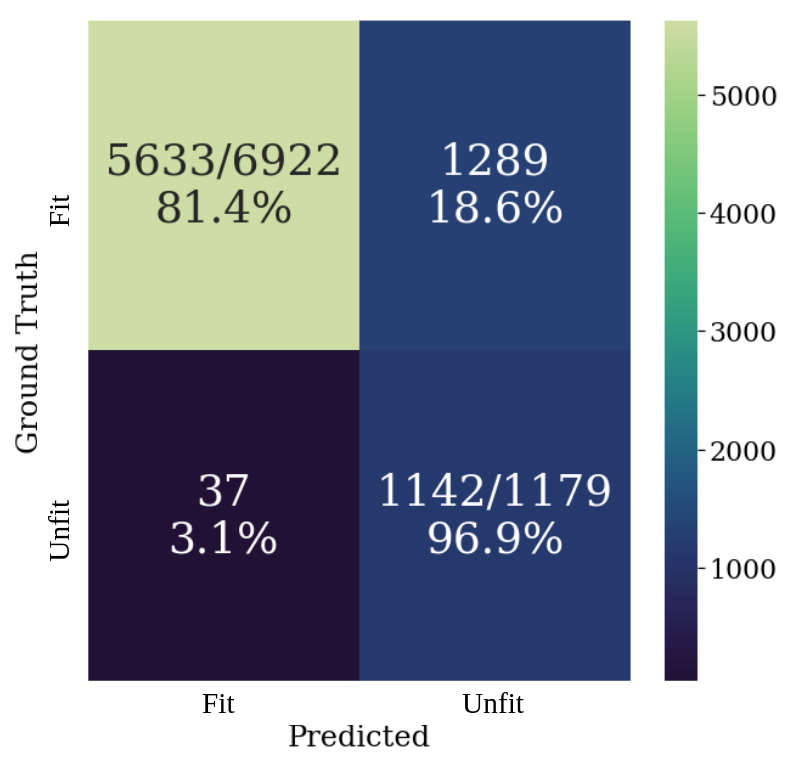} }}
        \subfloat[Multi-Class Confusion Matrix \label{subfig:multiclass}]{{\includegraphics[width=0.32\linewidth]{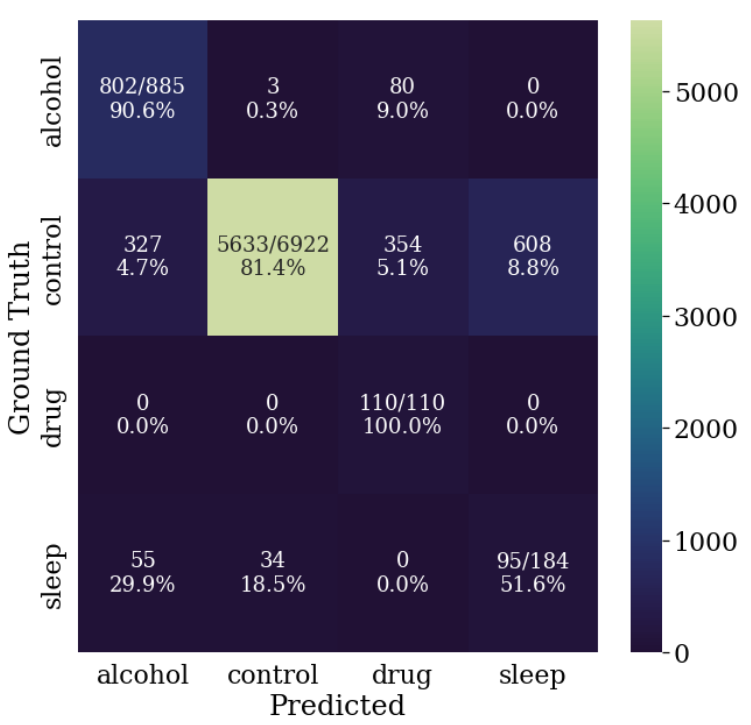} }}%
        \subfloat[DET Curve \label{subfig:det}]{{\includegraphics[width=0.32\linewidth]{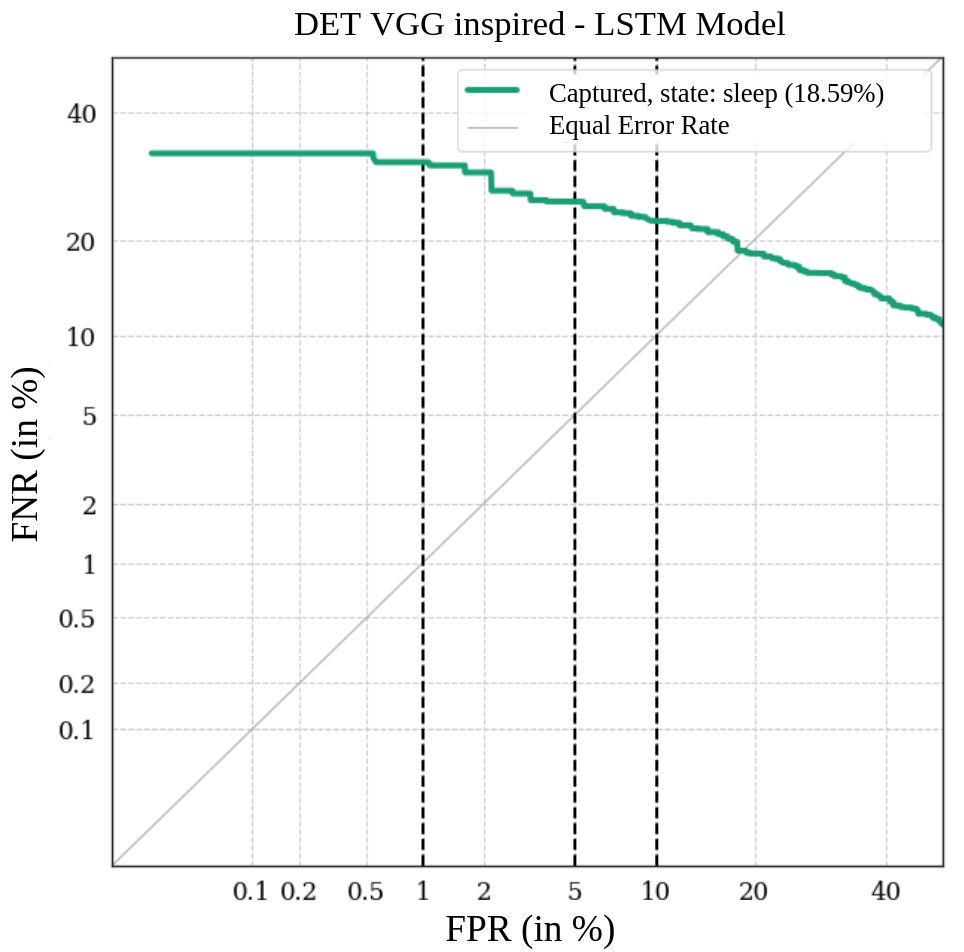} }}
    \end{center}
    \caption{Confusion Matrices and DET curve evaluated in the test set. The Models were trained using control subjects versus alcohol, drug consumption and sleepiness. The EER for the curve is shown in parentheses. The black dashed lines indicate two operational points for FNR$_{10}$, FNR$_{20}$, and FNR$_{100}$.}%
    \label{fig:metrics}%
\end{figure*}

\subsection{Results on Test Set}
\label{sec:test}

It was possible to obtain different metrics that allowed us to define the model's performance. One of them is the Confusion Matrix (CM), which was used to visualise the total performance of the prediction. Our results in the test set show a fit precision of 81.4\% and unfit precision of 96.9\%, as seen in Figure \ref{subfig:binary}. This allows us to detect 8 from 10 subjects with high accuracy. Additionally, it was possible to obtain the Confusion Matrix by state. For alcohol, control drug, and sleepiness. The well-predicted sequences obtained 90.6 \%, 81.4\%, 100\%, and 51.6\% precision, respectively, shown in Figure \ref{subfig:multiclass}. In the fourth class matrix, the sleepiness condition reached only an accuracy of 51.56\% and an EER of 18.59\%. Further, 8.8\% of the control subjects were wrongly classified as sleep. Then, there is a high similarity between sleep and control images.

As we can see in the results, we still have space for improvement, especially in the sleepiness category. It is essential to highlight that no subjects belong purely on a single class. Despite the fact that the rigorous work creating the dataset \cite{tapia2022alcohol,tapia2022ffd,causa2022ffd}, much of the time, subjects under the alcohol effect also present a sleepiness condition that can be seen as a real condition in real-life situations. The same situation can be assumed for drug conditions, as the confusion matrix shows a $9.0\%$ relation between drugs and alcohol. Therefore, there is a high correlation between the variables. All the control subjects were categorised based on self-reported for each volunteer at the capture time; however, they could have omitted not having slept well or consuming psychotropic substances before the test. Nevertheless, this condition represents very well the real problem. 

Table \ref{tab:metrics}, shows the metrics reached for our best model evaluated on the worst-performing class, which was  sleepiness. The EER achieved 18.59\%. The FNR$_{10}$ reached 22.5\% and FNR$_{20}$ reached 25.3\%. Those operating points can also be seen in Figure \ref{subfig:det}. The operating point chosen in the Confusion Matrices of Figure \ref{fig:metrics} corresponds with the EER, with a threshold of 0.047585.

\begin{table}[H]
\scriptsize
\caption{Summary results of the best model.}
\centering
    \begin{tabular}{c c  }
    \hline
    \hline
        Metrics & Values   \\ 
    \hline
        EER           & 0.1858945489 \\ 
        th EER        & 0.047585 \\ 
        Threshold     & 0.047585 \\ 
        FPR         &   0.1847826087 \\
        FNR         &   0.1862178561 \\
        FNR 10      &   0.2252239237 \\
        FNR 20      &   0.252672638 \\
        FNR 100     &   0.3140710777 \\ 
        th FNR 10   &   0.119005 \\
        th FNR 20   &   0.18455 \\
        th FNR 100  &   0.46277 \\
    \hline
    \hline
    \end{tabular}
\label{tab:metrics}
\end{table}

\begin{figure}[]
\centering
\includegraphics[scale=0.22]{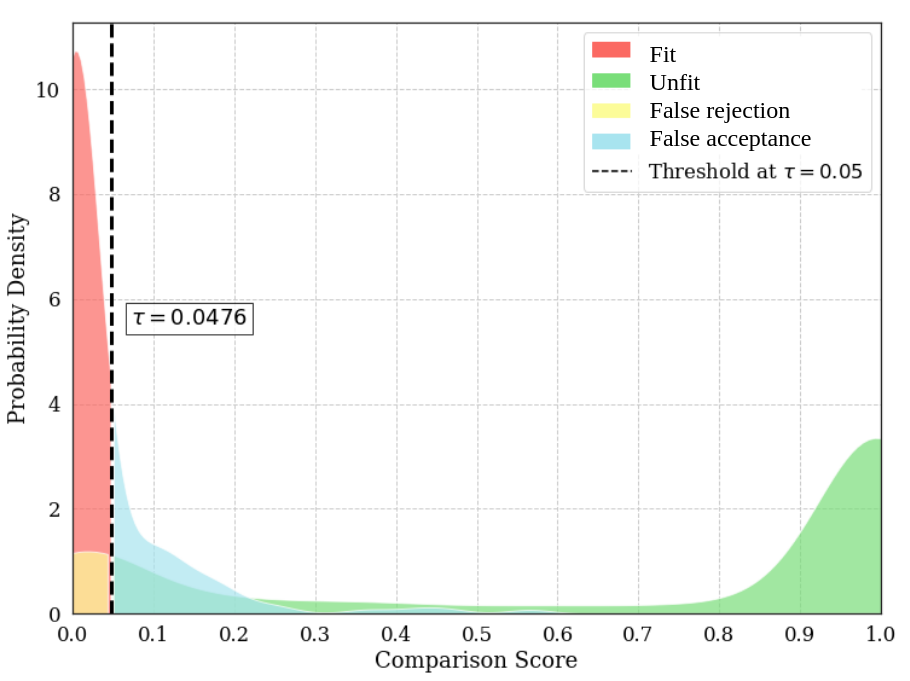}
\caption{Kernel Distribution Estimation Plot in linear scale.}
\label{fig:kde1}
\end{figure}

\begin{figure}[]
\centering
\includegraphics[scale=0.22]{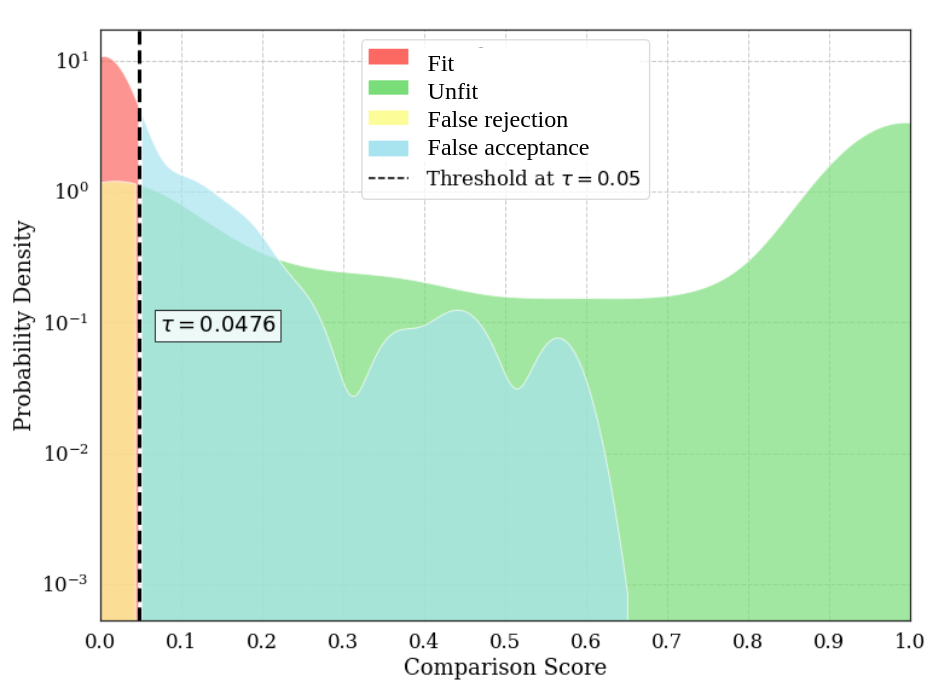}
\caption{Kernel Distribution Estimation Plot in logarithmic scale. Y-axis highlighted in relation to Figure \ref{fig:kde1}.}
\label{fig:kde2}
\end{figure}

Figure \ref{fig:kde1} shows the Kernel Distribution Estimation (KDE) Plot in linear scale, which depicts the probability density function of the continuous or non-parametric data variables. Figure \ref{fig:kde2} also shows a KDE plot but on a logarithmic scale to highlight the differences. Also, a black dot line is depicted, showing the best threshold for this system.

\subsection{Evaluation with other models}
\label{sec:comparison}

We compare our results with those obtained by Causa et al. \cite{causa2022ffd} using machine learning techniques and three classifiers Random Forest (FR), Gradient Boosting Machine (GBM), and Multi-Layer-Perceptron (MLP). As both works used iris sequences and the same database with the same train, test and validation partitions.

As shown in Table \ref{tab:comparison}, when analysing the overall accuracy of the models, our CNN-LSTM model obtains a value of 88.3\%, which is significantly higher than the methods compared (between 70.8\% and 75.5\%). Additionally, sensitivity, which provides the performance of the model to detect the states separately, is 88.0\% and 91.7\% for fit and unfit, respectively. This implies that our VGG16-inspired module can extract more relevant features than manual extraction \cite{causa2022ffd}. Additionally, in the next module, our LSTM uses information of only 8 frames, whereas Causa's method uses more than 75 to make an inference. Thus, the proposed LSTM is more optimal for determining the dependencies in time, which provides richer information.

\begin{table}[H]
\scriptsize
\centering
\caption{Comparison with state-of-art. Acc: Accuracy.}
\label{tab:comparison}
\begin{tabular}{cccccc}
\hline
Method & \multicolumn{1}{l}{Cond} & \begin{tabular}[c]{@{}c@{}}Sensitivity\\ (\%)\end{tabular} & \begin{tabular}[c]{@{}c@{}}Specificity\\ (\%)\end{tabular} & \begin{tabular}[c]{@{}c@{}}F1-Score\\ (\%)\end{tabular} & \begin{tabular}[c]{@{}c@{}}Accuracy\\ (\%)\end{tabular} \\ \hline
\multirow{2}{*}{RF \cite{causa2022ffd}} & Fit & 70.1 & 94.7 & 80.5 & \multirow{2}{*}{70.8} \\ \cline{2-5}
 & Unfit & 75.2 & 28.5 & 41.3 &  \\ \hline
\multirow{2}{*}{GMB \cite{causa2022ffd}} & Fit & 73.1 & 95.8 & 82.9 & \multirow{2}{*}{73.1} \\ \cline{2-5}
 & Unfit & 79.8 & 40.0 & 45.7 &  \\ \hline
\multirow{2}{*}{MLP \cite{causa2022ffd}} & Fit & 75.3 & 95.4 & 84.2 & \multirow{2}{*}{75.3} \\ \cline{2-5}
 & Unfit & 77.1 & 33.1 & 46.3 &  \\ \hline
\multirow{2}{*}{\textbf{\begin{tabular}[c]{@{}c@{}}CNN-LSTM\\ (Ours)\end{tabular}}} & Fit & 81.4 & 99.3 & 89.5 & \multirow{2}{*}{\textbf{83.6}} \\ \cline{2-5}
 & Unfit & 96.9 & 46.9 & 63.3 &  \\ \hline
\end{tabular}
\end{table}


\section{Conclusion}
\label{sec:conclusion}

This work shows that it is feasible to classify FFD using a stream of NIR images. The time-sequence information also is relevant to realise a more realistic prediction that can take advantage of the extraction of features when it is put in context with the LSTM network. We achieved state-of-the-art accuracy by extracting information from 8 frames, organised as a labelled $5D$ tensor. 
This can be reflected when comparing the accuracy and sensitivity of our model with previous works. 
The FFD-NIR-Seq is a dataset showing much of the reality, which is that subjects might present more than one state at a time. This assumption entails explaining the sleepiness state, which can be identified as a factor more complex to be predicted. 
In future work, we will continue to develop more precise methods and capture more images to improve the sleepiness class and the other categories. A lightweight version model with a reduced number of parameters should be developed in order to be implemented in mobile and portable devices. Additionally, data augmentation suitable for sequences should be implemented to improve the model's generalisation capabilities.

Finally, this work shows that there is still excellent applicability of our model in the real world when considering that from a pool of people that would try to perform their job in shift requires social and financial responsibility to save lives, only 81.4\% of them would be actually in an adequate condition to perform. Our CNN-LSTM network may help to prevent accidents, improve productivity and keep a better workspace.



\bibliographystyle{IEEEtran}
\bibliography{references.bib}

\end{document}